\documentclass[conference]{IEEEtran}
\IEEEoverridecommandlockouts
\usepackage{cite}
\usepackage{amsmath,amssymb,amsfonts}
\usepackage{algorithmic}
\usepackage{graphicx}
\usepackage{textcomp}
\usepackage{xcolor}
\usepackage{url}
\usepackage{graphicx}
\usepackage{subcaption}
\usepackage{booktabs}

\def\BibTeX{{\rm B\kern-.05em{\sc i\kern-.025em b}\kern-.08em
    T\kern-.1667em\lower.7ex\hbox{E}\kern-.125emX}}
\begin{document}

\title{Improving SCGAN's Similarity Constraint and Learning a Better Disentangled Representation\\
}

\author{\IEEEauthorblockN{1\textsuperscript{st} Iman Yazdanpanah}
\IEEEauthorblockA{\textit{Department of Electrical and Computer Engineering} \\
\textit{Isfahan University of Technology}\\
Isfahan, Iran \\
i.yazdanpanah@ec.iut.ac.ir}
\and
\IEEEauthorblockN{2\textsuperscript{nd} Ali Eslamian}
\IEEEauthorblockA{\textit{Department of Electrical and Computer Engineering} \\
\textit{Isfahan University of Technology}\\
Isfahan, Iran \\
a.eslamian@ec.iut.ac.ir}

}

\maketitle

\begin{abstract}
SCGAN adds a similarity constraint between generated images and conditions as a regularization term on generative adversarial networks. Similarity constraint works as a tutor to instruct the generator network to comprehend the difference in representations based on conditions. We understand how SCGAN works on a deeper level. This understanding makes us realize that the similarity constraint functions like the contrastive loss function. Two major changes we applied to SCGAN to make a modified model are using SSIM to measure similarity between images and applying contrastive loss principles to the similarity constraint. The modified model performs better using FID and Factor metrics. On the MNIST and Fashion-MNIST datasets, the modified model achieves log-likelihood values of 234.8 and 332.6, respectively, surpassing SCGAN’s 232.5 and 324.2. The modified model exhibits improvements in FID, attaining values of 3.42 and 12.97 for MNIST and Fashion-MNIST, respectively, compared to SCGAN’s 4.11 and 14.63. In terms of disentanglement, the modified model shows clear advantages, achieving values of 0.89 and 0.91 for MNIST and Fashion-MNIST, respectively, compared to SCGAN’s 0.77 and 0.89. These improvements show that the modified model has effectively learned a more disentangled representation compared to SCGAN. The modified model also has better generalisability compared to other models. \footnote{The code is available at \\
\url{https://github.com/Iman-yazdanpanah/contrastive-SSIM-SCGAN}.}
\end{abstract}

\section{Introduction}
The focus of representation learning is to describe training set observations in a low-dimensional latent space. This allows us to learn a mapping function that can transform a point in the latent space to a point in the original domain. Each point in the latent space represents a high-dimensional image \cite{b1}. The key advantage of representation learning is its ability to determine the most important features for describing observations and how to generate these features from raw data.

Representations can be categorized as disentangled or distributed. Disentangled representations capture and separate the underlying factors of variation in data, while distributed representations encode these factors in a distributed manner. Altering a unit in latent space causes a corresponding alteration in a generative factor within a disentangled representation. A significant amount of research has been dedicated to representation learning, with generative modeling playing a crucial role\cite{b2, b3}. The model’s ability to generate realistic samples suggests a deeper understanding of the underlying representation. Variational Auto Encoder (VAE) \cite{b4} and generative adversarial networks (GAN) \cite{b5} are two generative models that can learn a disentangled representation. GAN learns a distributed representation, but Conditional GAN (CGAN) \cite{b6} learns a disentangled representation in a supervised manner. Information-maximizing GAN (InfoGAN) \cite{b7-chen2016infogan} and Similarity Constraint GAN (SCGAN) \cite{b8-li2018scgan} do so in an unsupervised manner. We will discuss each model in section \ref{sec3:background}.  

In \cite{b3}, it has been shown that unsupervised disentangled representation learning without inductive biases is theoretically impossible. Existing inductive biases and unsupervised methods do not allow consistent learning of disentangled representations. Any regularization term is considered an inductive bias \cite{b9-goyal2022inductive}. Therefore, SCGAN uses inductive biases in its similarity constraint (SC). A superior model should perceive image similarity as humans do, based on the structure of images. So we can use SSIM \cite{b10-wang2004image} as a similarity measurement. SSIM compares images based on structure, contrast, and luminance. By using a structural similarity measurement, we believe that the modified model learns an interpretable representation.

In similarity learning \cite{b11-chopra2005learning, b12-taigman2014deepface, b13-schroff2015facenet}, contrastive loss works just like similarity constraint in SCGAN. The model aims to learn a representation where positive pairs (images from the same class) are closer together in the embedding space than negative pairs (images from different classes). SC aims to create similar images from the same latent code and dissimilar images from different latent codes. We study contrastive loss and determine that the SC in SCGAN is not efficiently activated. Because SC applies to 32 images, calculations for images with different latent code $c$ are more effective. In contrastive loss, the number of negative pairs should be close to negative pairs, or else the model learns to either place all the images far from each other (when the number of negative pairs is much bigger than the number of positive pairs) or close to each other (when the number of positive pairs is much bigger than the number of negative pairs). Therefore, we should equalize the contribution of images with the same and different latent codes $c$ on the SC.

The evaluation method employed in \cite{b8-li2018scgan} cannot assess the model’s capabilities. To evaluate SCGAN’s performance, the authors calculated the log-likelihood estimation, which statistically compares synthetic images with test data. A model can suffer from mode collapse and still have a high log-likelihood estimation. So we use FID \cite{b14-heusel2017gans}, which compares both the quality and diversity of generated images. There are numerous ways to measure the disentanglement of learned representation \cite{b3,b15-kim2018disentangling,b16-eastwood2018framework}, but we use FactorVAE \cite{b15-kim2018disentangling}.

Our main contributions are:  

\begin{itemize}

\item

Using SSIM to measure similarity between synthetic images so that the modified model is more interpretable.

\item 

Improve SCGAN’s similarity constraint by trying to equalize the effect of each term. We achieve this goal by reducing the calculations for images of different $c$ and changing the functions used in each term of the equations.

\item

Using FID and FactorVAE to measure the modified model’s performance on different datasets.

\end{itemize}

In the rest of this paper, we begin by reviewing the related works in section \ref{sec2:Related Work}. In section \ref{sec3:background}, we review GAN, CGAN, and InfoGAN, then thoroughly review SCGAN and understand how SCGAN works on a deeper level. A thorough understanding of SCGAN’s mechanisms and limitations is crucial for devising effective ways to enhance its performance. In section \ref{sec4:Modified model}, we apply some changes to SCGAN’s similarity constraint and we introduce the modified model. In section \ref{sec5:Experiments}, we implement the modified model and do experiments on datasets such as MNIST, Fashion-MNIST, CELEBA, and CIFAR10. then, we compare the modified model’s performance with other models. Finally, in section \ref{sec6:Conclusion}, we will conclude and discuss future work ahead.

\section{Related Work}\label{sec2:Related Work}
Several research studies have investigated distributed representation and disentangled representation. VAE \cite{b4} and GAN \cite{b5} have recently seen a lot of interest in generative modeling problems. In both approaches, a deep neural network is trained as a generative model by using backpropagation, enabling the generating of new images without explicitly learning the underlying data distribution. VAE maximizes a lower bound on the marginal likelihood, which is expected to be tight for accurate modeling \cite{b17-sonderby2016ladder,b18-paige2017learning}. However, GAN optimizes a minimax game function via a discriminative adversary. VAE and GAN learn distributed representation. 

CGAN \cite{b6}, InfoGAN \cite{b7-chen2016infogan}, and SCGAN \cite{b8-li2018scgan} are all capable of learning disentangled representations. While CGAN uses a supervised method in its training process, InfoGAN and SCGAN utilize unsupervised methods. CGAN learns by incorporating additional information $c$, whereas InfoGAN ensures that the generator does not ignore $c$ by maximizing mutual information between $c$ and synthetic images. Furthermore, InfoGAN employs an additional network $Q$ to maximize the variational lower bound of mutual information. On the other hand, SCGAN endeavors to learn disentangled representation by introducing a similarity constraint between the latent vector $c$ and synthetic images so that images generated with the same latent vector exhibit visual similarities. Other works \cite{b3, b15-kim2018disentangling, b19-chen2018isolating, b20-higgins2016beta} employ VAE models for acquiring disentangled representation.

SCGAN's similarity constraint uses Euclidean distance to measure the similarity between generated images. Different distance metrics such as Gaussian radial basis function, cosine distance, Manhattan distance, and others can also be utilized. In disentangled representation learning, it is often assumed that the learned representation should be human-interpretable \cite{b21-stammer2022interactive}. This allows us to employ inductive biases more effectively. Therefore, we utilize SSIM \cite{b10-wang2004image} for measuring the similarity between produced images based on their structure, contrast and luminance. The modified model is more interpretable.

SCGAN’s authors implemented Gaussian Parzen window log-likelihood estimation to measure the performance. Models can suffer from mode collapse and still have a high log-likelihood. We use FID \cite{b14-heusel2017gans} which considers both quality and diversity of generated images. If a model suffers from mode collapse, the FID score will be worse because the diversity of generated images will decrease. Inception score \cite{b22-salimans2016improved} is another metric that works like FID. In \cite{b8-li2018scgan}, the authors do not employ any quantitative metrics to evaluate the disentanglement of the learned representation. There are several metrics to measure disentanglement \cite{b15-kim2018disentangling, b16-eastwood2018framework, b20-higgins2016beta}.

In similarity learning, the goal is to learn a mapping that maps input images into a target space such that the norm in the target space approximates the “semantic” distance in the input space, meaning similar images are mapped near each other in the target space and dissimilar images are mapped far from each other. Reference \cite{b11-chopra2005learning} uses a contrastive loss function and \cite{b13-schroff2015facenet} employs a triplet loss function to achieve a good similarity learning objective. In similarity learning, the number of negative pairs is usually much more than the number of positive pairs, which is considered a problem \cite{b23-pan2021contrastive}, so researchers tried to use generated images only as positive pairs \cite{b24-khosla2020supervised}.

\section{Background}\label{sec3:background}
In this section, we review generative models such as GAN, CGAN, InfoGAN and SCGAN.

\subsection{GAN}\label{GAN}
GAN was introduced as a new way to train generative models. It comprises two networks, a generator network $G$ and a discriminator network $D$ and they interact with each other in an adversarial manner. The goal is to learn a generator distribution $P_G$ that matches the real data distribution $P_{data}$. $G$ maps noise $z$ drawn from $P_{noise}$ to sample $G(z)$. $D$ tries to discriminate real data from generated images $x$ by $G$.

GAN learns a distributed representation. The objective function between generator $G$ and discriminator $D$ is a minimax game, and it is given as follows:

\begin{equation}
\begin{split}
\min_{G} \max_{D} V(D, G) = & \operatorname{E}_{x \sim P_{\text{data}}} \left[ \log D(x) \right] \\
& + \operatorname{E}_{z \sim P_z} \left[ \log(1 - D(G(z))) \right]
\end{split}
\end{equation}

\subsection{CGAN}\label{CGAN}
CGAN \cite{b6} uses additional information $c$ as input to both generator and discriminator, thus their outputs are $G(z|c)$ and $D(x|c)$ respectively. CGAN learns a disentangled representation in a supervised manner. The objective of CGAN is as follows:

\begin{equation}
\begin{split}
\min_{G} \max_{D} V(D, G) = & \operatorname{E}_{x \sim P_{\text{data}}} \left[ \log(D(x|c))) \right] \\
& + \operatorname{E}_{z \sim P_{noise}} \left[ \log(1 - D(G(z|c))) \right] \\
\end{split}
\end{equation}

\subsection{InfoGAN}
In order to learn a disentangled representation, InfoGAN \cite{b7-chen2016infogan} proposes a regularizer based on mutual information. As the goal is not to disentangle all latent codes but to disentangle a subset, InfoGAN splits the latent code into two parts, the disentangled code $c$ and the remaining code $z$ which provides more randomness. InfoGAN learns a disentangled representation in an unsupervised manner by maximizing mutual information between $c$ and generated images $G(z,c)$ but maximizing the mutual information $I (c;G(z,c))$ directly is a difficult task, so InfoGAN uses a variational lower bound $L_I(G, Q)$ by the technique known as Variational Information Maximization \cite{b25-barber2004algorithm}.
They use additional network $Q(c|x)$ to approximate $P(c|x)$. Then, the lower bound is easily approximated and used as a regularization term with a hyperparameter $\lambda$ in the objective as follows:
\begin{equation}
    \min_{G,Q} \max_{D} V(D, G) - \lambda L_I(G, Q)
\end{equation}

\subsection{SCGAN} \label{SCGAN}
SCGAN \cite{b8-li2018scgan} learns a disentangled representation in an unsupervised manner by using a similarity constraint as a regularizer, which is a function of latent vector $c$. In order to define SC, they had to define similarity $Sim(x_i, x_j)$ on a pair of images $x_i, x_j$ which measures the difference between $x_i$ and $x_j$. $Sim$ can be any function but it must follow these conditions:
\begin{itemize}
\item
$Sim$ must satisfy the smoothness assumption so when $z$ or $c$ changes smoothly, similarity also changes smoothly.
\item 
If $x_i$ and $x_j$ are similar on the measure space, $Sim(x_i, x_j)$ should be small, otherwise $Sim(x_i, x_j)$ should be large.
\item
Similarly is symmetrical, meaning that $Sim(x_i, x_j) = Sim(x_j, x_i)$.
\end{itemize}

Euclidean distance satisfies these conditions so SCGAN uses it to measure similarity between generated images $G(z,c)$. The latent vector, denoted as $c$, can either be continuous or discrete. The discrete conditional variable is utilized to capture the main differences in real data, such as class labels. On the other hand, the continuous conditional variable is used to capture slowly changing attributes like rotation and size of objects. For discrete conditional variables, SC is defined as follows:
\begin{equation} \label{eq4}
\begin{aligned}
\text{SC}(X,c) &= \frac{1}{N(N-1)} \sum_{i} \sum_{j \neq i} \\
& \left( \langle c_i, c_j \rangle \, Sim(x_i, x_j) + \frac{1-\langle c_i, c_j \rangle}{Sim(x_i, x_j)} \right)
\end{aligned}
\end{equation}

$N$ is the batch-size of $x$ and $c$ and it is equal to 32 and $\langle .,. \rangle$ denotes the inner product. For continuous conditional variables, SC is defined as follows:

\begin{equation}\label{eq5}
\begin{aligned}
\text{SC}(X,c) &= \frac{1}{N(N-1)} \sum_{i} \sum_{j \neq i} \\
& \left(1 - \lvert c_i, c_j \rvert \right) Sim(x_i, x_j) + \frac{\lvert c_i, c_j \rvert}{Sim(x_i, x_j)}
\end{aligned}
\end{equation}

Operator $\lvert . \rvert$  denotes absolute value. 

The objective of SCGAN is then defined as follows:
\begin{equation}
    \min_{G} \max_{D} V(D, G) - \lambda \text{SC}(x, c)
\end{equation}

$\lambda$ is a hyperparameter. SC is a function of $c$ and not $z$ therefore, minimizing SC is only related to $c$. So its goal is that generated images with the same $c$ should be similar and generated images with different $c$ should be dissimilar. In the discrete case in \ref{eq4}, if $x_i$ and $x_j$ are from different classes then $\langle c_i,c_j \rangle = 0$ and only the term $\frac{1}{Sim(x_i,x_j)}$ contributes to SC. In this case, minimizing SC is equivalent to maximizing dissimilarity between $x_i$ and $x_j$. In contrast, if $x_i$ and $x_j$ are from the same classes then $\langle c_i,c_j \rangle = 1$ and only the term  $Sim(x_i,x_j)$ contributes to SC. In this case, minimizing SC is equivalent to maximizing the similarity between $x_i$ and $x_j$. The continuous SC in \ref{eq5} functions the same. 

In the discrete case, the batch-size is 32, so there are 32 generated images that contribute to SC. On average, the term $Sim(x_i,x_j)$ contributes to SC about 32 times and the term $\frac{1}{Sim(x_i,x_j)}$ contributes to SC about 464 times. SC is functioning like contrastive loss, so both terms in SC should contribute at the same level. In section \ref{sec4:Modified model}, we will introduce the modified model.

\section{Modified Model}\label{sec4:Modified model}
In this section, we investigate changes that can be applied to SC and improve the model's performance.

\subsection{Structural Similarity}
Euclidean distance has some drawbacks. It assumes that the variables are independent and have equal importance, which may not be true in some cases depending on the dataset on which we train the model. The Effectiveness of Euclidean distance as a similarity metric diminishes as the dimensionality of data increases. This has to do with the curse of dimensionality. So we should use another similarity metric to measure the similarity between images. We suggest using SSIM \cite{b10-wang2004image} to measure the similarity between synthetic images so that the modified model is more interpretable. SSIM compares images based on structure, contrast, and luminance.

\subsection{Similarity Constraint}
As mentioned in section \ref{SCGAN}, SC functions like contrastive loss and on average, the contribution ratio of images with the same classes to images with different classes is $\frac{464}{32} = 14.5$. This means that in \ref{eq4}, the term $Sim(x_i,x_j)$ has less effect in the training process and can be ignored, so the model cannot generate similar images with the same $c$ and only learns to generate dissimilar images. In this section, we improve the similarity of SCGAN constraint by trying to equalize the effect of each term in \ref{eq4}. 

We attempted to minimize the influence of $\frac{1}{Sim(x_i,x_j)}$ on SC. This was accomplished by selecting 10 and 18 images at random from a collection of 32 generated images. We achieved this by randomly choosing 10 and 18 images from 32 generated images. We performed a series of experiments to determine the optimal number of images to choose, and we found that 10 and 18 images produced the best results. These numbers can be considered a hyperparameter of our approach. In this case, on average the term $\frac{1}{Sim(x_i,x_j)}$ contributes to SC 88 times and the term $Sim(x_i,x_j)$ contributes to SC 28 times. With this simple change, the contribution ratio of images with the same classes to images with different classes is now $\frac{88}{28}=3.14$ which means that the modified model will not ignore $Sim(x_i,x_j)$ during training and the modified model can learn to generate similar images. 

To have a better contrastive SC, we tried different functions, In other words, Instead of the conventional $\left(\text{Sim}, \frac{1}{\text{Sim}} \right)$, we conducted experiments with alternatives such as $\left(\text{Sim}^2, \frac{1}{\text{Sim}^2} \right)$, $\left(e^{\text{Sim}},\frac{1}{e^{\text{Sim}}} \right)$, and others. The optimal result was achieved using $\left(e^{\text{Sim}}, \frac{1}{e^{\text{Sim}}} \right)$.

The modified model's SC for discrete conditional variables is as follows:
\begin{equation}
\begin{aligned}
SC(x, c) &= \frac{1}{N_1 \times N_2} \sum_{i \in \mathbb{N}_1} \sum_{j \in \mathbb{N}_2, j > i} \\
& \left( \lambda_1 \langle c_i, c_j \rangle e^{\text{SSIM}(x_i, x_j)} + \lambda_2 \langle c_i, c_j \rangle e^{-\text{SSIM}(x_i, x_j)} \right)
\end{aligned}
\end{equation}

$\lambda_1$ and $\lambda_2$ are hyperparameters and we tune them by conducting experiments. We found out that the model performs better with $\lambda_1 = e$  and $\lambda_2 = e^{1.5}$ . $\lambda_1$ and $\lambda_2$ help us to have more power in equalizing the effect of each term in SC. For continuous conditional variables, we apply the same changes to \ref{eq5}.

Hence, the modified model is defined as the following minimax game with a similarity constraint regularizer:
\begin{equation}
    \min_{G} \max_{D} V(D, G) - \text{SC}(x, c) 
\end{equation}

\section{IMPLEMENTATION AND EXPERIMENTS}\label{sec5:Experiments}
The modified model’s architecture and SCGAN’s architecture are the same. SCGAN and modified model only differ in their similarity constraint, which has nothing to do with their architecture. Their implementation details are similar, and further implementation information can be found in \cite{b8-li2018scgan}. InfoGAN introduces extra network $Q$, which means its architecture is more complicated and requires more parameters to train. 

We train the modified model on MNIST \cite{b26-lecun1998gradient}, Fashion-MNIST \cite{b27-xiao2017fashion}, CELEBA \cite{b28-liu2015deep} and CIFAR10 \cite{b28-liu2015deep} and compare its performance to other models' performances. For all experiments, we use Adam optimizer \cite{b30-kingma2014adam} and batch-size is set to 32.

\subsection{MNIST and Fashion-MNIST}
The MNIST dataset \cite{b26-lecun1998gradient} is a widely used benchmark dataset in machine learning. It comprises 70,000 grayscale images of handwritten digits from 0 to 9, with each image being a $28 \times 28$ pixel square. MNIST is often used for tasks such as image classification and digit recognition, serving as a starting point for numerous beginners in the field because of its simplicity and ease of use.

Fashion-MNIST \cite{b27-xiao2017fashion} is another popular dataset commonly used for image classification tasks. Similar to MNIST, it contains 70,000 grayscale images, but instead of handwritten digits, it comprises 10 different categories of fashion items, such as T-shirts, dresses, sneakers, and more. Fashion-MNIST provides a more challenging task compared to MNIST, as it requires models to classify different clothing accurately.

Both datasets include ten classes which are likely to be captured by categorical conditional variables. We train CGAN, InfoGAN, SCGAN and the modified model on MNIST and Fashion-MNIST datasets respectively and use the Gaussian Parzen window to estimate the log-likelihood of each GAN. We use FID \cite{b14-heusel2017gans} and FactorVAE \cite{b15-kim2018disentangling} to measure each model's performance and disentanglement.

\begin{figure}
    \centering
    \begin{subfigure}{0.45\textwidth}
        \includegraphics[width=\linewidth]{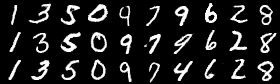}
        \caption{Samples from the model trained on MNIST}
    \end{subfigure}
    \hfill
    \begin{subfigure}{0.45\textwidth}
        \includegraphics[width=\linewidth]{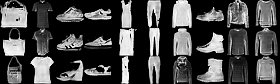}
        \caption{Samples from the model trained on Fashion-MNIST}
    \end{subfigure}
    \caption{Generated images by the modified model on Fashion-MNIST and MNIST. Images in each column have the same $c$ and are from one class.}
    \label{fig1}
\end{figure}

We train all models 25 epochs and understand that categorical conditional variables can capture class labels (e.g., digit type, clothing type). Generated images are shown in Fig. \ref{fig1}. Also, Table~\ref{tab:log_likelihood} illustrates the results of the Gaussian Parzen window log-likelihood estimate for the mentioned models on MNIST and Fashion-MNIST test data. The modified model has the greatest log-likelihood on test data which means the generated images’ distribution is closer to the real data’s distribution. The FID score for mentioned GANs is shown in Table~\ref{tab:fid_scores}. The best result has been achieved by the modified model. Hence, the images generated by the modified model showcase improved quality, heightened realism, and expanded diversity. 

\begin{table}
    \centering
    \caption{\textsc{Log-likelihood estimates for MNIST and Fashion-MNIST using Gaussian Parzen window. Numbers are mean log-likelihood with the standard error of the mean computed across examples}}
    \label{tab:log_likelihood}
    \begin{tabular}{lcc}
        \toprule
        \textbf{Model} & \textbf{MNIST} & \textbf{Fashion-MNIST} \\
        \midrule
        CGAN & $228.9 \pm 2.1$ & $311.8 \pm 2$ \\
        InfoGAN & $232 \pm 2.1$ & $313.5 \pm 2$ \\
        SCGAN & $232.5 \pm 2$ & $324.2 \pm 2$ \\
        \textbf{Modified Model} & $234.8 \pm 2.1$ & $332.6 \pm 1.9$ \\
        \bottomrule
    \end{tabular}
\end{table}

In terms of disentanglement, there are many metrics introduced in \cite{b3} like Modularity \cite{b31-ridgeway2018learning}, DCI \cite{b16-eastwood2018framework} and FactorVAE \cite{b15-kim2018disentangling}. We choose FactorVAE to measure each GAN's ability to learn a disentangled representation because it’s easier to implement and also provides a quantitative metric that measures disentanglement achieved by the model. This score allows for a more objective evaluation of the disentanglement quality compared to Modulariy and DCI, which rely on qualitative assessments. Table~\ref{tab:factorvae_scores} displays the FactorVAE score for each GAN on MNIST and Fashion-MNIST. On MNIST, CGAN and the modified model performed almost at the same level, although the modified model performed slightly better.

On Fashion-MNIST, SCGAN and the modified model performed almost at the same level, although the modified model performed slightly better. Table~\ref{tab:factorvae_scores} shows that CGAN, InfoGAN and SCGAN learned a good disentangled representation only on one dataset but the modified model learned a good disentangled representation on both datasets, which shows that the modified model has better generalisability than all other models.

\subsection{CELEBA}
The CelebA \cite{b28-liu2015deep} dataset is a large-scale face attributes dataset that contains over 200,000 celebrity images. It is widely used for tasks such as face recognition, facial attribute analysis, and face synthesis. Each image in CelebA is annotated with $40$ attribute labels, including gender, hair color, presence of glasses, and more. This dataset offers a diverse range of facial features and poses, making it valuable for training models to understand and analyze human faces. Unlike MNIST and Fashion-MNIST, CELEBA does not have test data so we cannot evaluate models' performances quantitatively on CELEBA. We can conduct experiments and compare models qualitatively.

\begin{figure*}
    \centering
    \begin{subfigure}{0.45\textwidth}
        \includegraphics[width=\linewidth]{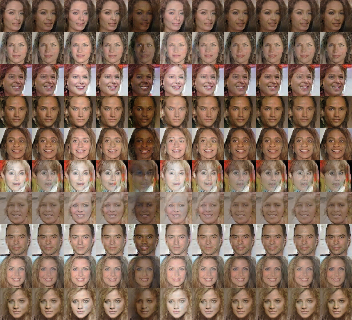}
        \caption{}
    \end{subfigure}
    \hfill
    \begin{subfigure}{0.45\textwidth}
        \includegraphics[width=\linewidth]{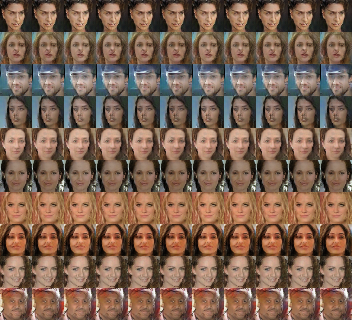}
        \caption{}
    \end{subfigure}
    \caption{Generated samples on CELEBA by (a) the modified model and (b) SCGAN. Each column is sampled from different $z$ while fixing $c_i$. Each row is sampled from different $c_i$ while fixing $z$}
    \label{fig2}
\end{figure*}

We trained both SCGAN and the modified model on CELEBA. In Fig. \ref{fig2}, generated images by both models are shown. Images in the same row have a fixed $z$ and images in the same column have a fixed $c$. This way, we can see if any factor of variation is learned by $c$. Images generated by SCGAN in a row just look the same. Images generated by the modified model in a row differ in features, so the model learned a better disentangled representation. For example, in column 5 of \ref{fig2}, the race of the people is changed.

\begin{table}
    \centering
    \caption{\textsc{FID scores \cite{b14-heusel2017gans} for MNIST and Fashion-MNIST. A lower FID indicates better quality and diversity of images}}
    \label{tab:fid_scores}
    \begin{tabular}{lcc}
        \toprule
        \textbf{Model} & \textbf{MNIST} & \textbf{Fashion-MNIST} \\
        \midrule
        CGAN & 5.04 & 15.51 \\
        InfoGAN & 4.60 & 17.65 \\
        SCGAN & 4.11 & 14.63 \\
        \textbf{Modified Mode} & 3.42 & 12.97 \\
        \bottomrule
    \end{tabular}
\end{table}

\begin{table}
    \centering
    \caption{\textsc{FactorVAE scores \cite{b15-kim2018disentangling} to measure disentanglement. The modified model has better generalizability}}
    \label{tab:factorvae_scores}
    \begin{tabular}{lcc}
        \toprule
        \textbf{Model} & \textbf{MNIST} & \textbf{Fashion-MNIST} \\
        \midrule
        CGAN & 0.88 & 0.72 \\
        InfoGAN & 0.83 & 0.75 \\
        SCGAN & 0.77 & 0.89 \\
        \textbf{Modified Model} & 0.89 & 0.91 \\
        \bottomrule
    \end{tabular}
\end{table}

\subsection{CIFAR10}
CIFAR-10 \cite{b29-krizhevsky2009learning} is a well-known dataset used for image classification tasks. It comprises 60,000 color images, divided into ten classes such as airplanes, cars, birds, cats, and more. Each image in CIFAR-10 has a resolution of $32 \times 32$ pixels, providing a more complex and detailed dataset compared to MNIST and Fashion-MNIST. CIFAR-10 is often used to evaluate the performance of models in handling real-world color images and their ability to classify objects accurately. Like CELEBA, CIFAR10 does not have test data, so we cannot evaluate models’ performances quantitatively. We can conduct experiments and compare models qualitatively.

\begin{figure*}
    \centering
    \includegraphics[width=0.95\linewidth, height=7cm]{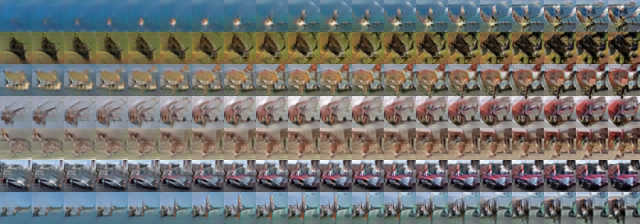}
    \caption{Generated samples on CIFAR10. Each row is sampled from different $c_1$ (left to middle images) and $c_2$ (middle to right images) while fixing $z$. The size of objects changes from left to middle, and color of objects and background changes from middle to right.}
    \label{fig:figure3}
\end{figure*}

In \cite{b8-li2018scgan}, the authors claimed SCGAN learned attributes like the size of objects by a continuous conditional variable and learned to change the color of objects and background by another continuous conditional variable $c$. We show in Fig. \ref{fig:figure3} that the modified model has the same ability.  
\subsection{Time Consumption}
InfoGAN and CGAN require less time to train. SCGAN takes about five times more time to be trained because of all the calculations in SC.

The modified model requires fewer calculations. On average, the pair-wise calculations for the modified model are reduced from 496 to 116. SSIM calculation takes more time than Euclidean distance calculation. The training time for SCGAN and the modified model is almost equal, but the latter is slightly faster to train. Table~\ref{tab:training_duration} displays the average time required for completing one step of training SCGAN and the modified model on each dataset. The modified model shows a slight increase in speed on MNIST and CIFAR10 datasets, approximately twice as fast as SCGAN on CELEBA, and equally fast compared to SCGAN on Fashion-MNIST.

\begin{table}
    \centering
    \caption{\textsc{The average duration for one step of training SCGAN and the modified model on different datasets}}
    \label{tab:training_duration}
    \begin{tabular}{lccc}
        \toprule
        \text{Conditional Variable} & \text{Dataset} & \text{SCGAN} & \textbf{Modified Model} \\
        \midrule
        Discrete & MNIST & 0.2s & 0.19 \\
                  & Fashion-MNIST & 0.22s & 0.22 \\
                  & CELEBA & 0.61s & 0.33 \\
        \midrule
        Continuous & CIFAR10 & 0.35s & 0.27 \\
        \bottomrule
    \end{tabular}
\end{table}

\section{Conclusion}\label{sec6:Conclusion}
In this work, we understand how SCGAN works on a deeper level. This understanding makes us realize that the similarity constraint is used as a regularizer and functions like the contrastive loss function. We improved SCGAN's similarity constraint by using SSIM to measure the similarity between synthetic images and leveling the contribution of each term of the similarity constraint to have a more contrastive regularization term. Experiments show that our proposed model performs better and generates more realistic and diverse images based on the FID score. The modified model has a better generalisability and has learned a better disentangled representation of MNIST and Fashion-MNSIT. On CELEBA and CIFAR10, it is visible that the modified model learns a better disentangled representation.

We believe calculating SSIM takes much time and reducing the calculations of similarity constraint is not that much effective. The modified model is still 4-5 times slower than CGAN and InfoGAN. For future work, we suggest using other structure-based similarity criteria that are faster to calculate. For instance, fine-tuning the pre-trained VGG-16 \cite{b32-simonyan2014very} as a perceptual loss and applying it to each dataset \cite{b33-johnson2016perceptual}.

We randomly chose 10 and 18 generated images and applied the similarity constraint to them, which improved the contribution ratio. By doing so, we reduced the number of times that negative pairs contributed to SC. For future work; we suggest increasing the number of times that the positive pairs contribute to SC. Moreover; we suggest saving generated images and using them only as positive pairs in the next step of the training procedure. 

\bibliographystyle{unsrt}
\bibliography{references.bib}
\end{document}